# AI-Driven anemia diagnosis: A review of advanced models and techniques

Abdullah Al Mahmud [a], Prangon Chowdhury [b, *], Mohammed Borhan Uddin [a], Khaled Eabne Delowar [a], Tausifur Rahman Talha [a], Bijoy Dewanjee [a]

[a] Department of Computer Science and Engineering, International Islamic University Chittagong (IIUC), Chittagong 4318, Bangladesh

[b] Department of Electrical and Electronic Engineering, Ahsanullah University of Science and Technology (AUST), Dhaka 1208, Bangladesh

* Corresponding author: Email address: prangon21@gmail.com

**Abstract**

Anemia, a condition marked by insufficient levels of red blood cells or hemoglobin, remains a widespread health issue affecting millions of individuals globally. Accurate and timely diagnosis is essential for effective management and treatment of anemia. In recent years, there has been a growing interest in the use of artificial intelligence techniques i.e. machine learning (ML) and deep learning (DL) for the detection, classification, and diagnosis of anemia. This paper provides a systematic review of the recent advancements in this field, with a focus on various models applied to anemia detection. The review also compares these models based on several performance metrics, including accuracy, sensitivity, specificity, and precision. By analyzing these metrics, the paper evaluates the strengths and limitation of discussed models in detecting and classifying anemia, emphasizing the importance of addressing these factors to improve diagnostic accuracy.

## 1. Introduction

Anemia is a significant global health issue, characterized by a reduction in the number of red blood cells (RBCs) or a decrease in hemoglobin concentration, impairing the body's ability to transport oxygen efficiently. The World Health Organization (WHO) defines anemia as a hemoglobin concentration below a specific threshold, which varies by age, gender, physiological status, smoking habits, and altitude. According to WHO, anemia is diagnosed at hemoglobin levels below 110 g/L for children under 5 and pregnant women, and below 120 g/L for non-pregnant women at sea level [Anaemia (who.int)]. The condition has various causes, including nutritional deficiencies, chronic diseases, genetic disorders, and environmental factors, affecting approximately 1.6 billion people worldwide, with the highest prevalence in low- and middle-income countries. If left untreated, anemia can lead to severe complications, particularly in vulnerable groups such as children, pregnant women, and the elderly. Effectively addressing anemia requires an understanding of its causes and the implementation of prevention and treatment strategies tailored to the needs of different populations [1],[2].

There are several types of anemia, each with distinct causes and characteristics. The most common is iron deficiency anemia (IDA), resulting from inadequate iron intake, malabsorption, or chronic blood loss. IDA poses a significant risk to women of reproductive age and children, affecting an estimated 29.9% of the global population [ https://doi.org/10.1016/j.reprotox.2023.108381 ]. Women of reproductive age are particularly vulnerable due to increased iron demands during pregnancy, lactation, menstrual bleeding, and nutritional deficiencies throughout their reproductive cycle [3]. Another common type is megaloblastic anemia, primarily caused by vitamin B12 or folate deficiency, leading to the production of abnormally large RBCs [4]. Additionally, hemolytic anemia occurs when RBCs are destroyed faster than they can be produced, which can result from autoimmune disorders, infections, or inherited conditions such as sickle cell disease and thalassemia [5].

The causes of anemia can be broadly classified into three mechanisms: blood loss, decreased RBC production, and increased RBC destruction. Nutritional deficiencies, particularly in iron, vitamin B12, and folate, are major contributors to anemia globally. Chronic diseases, such as chronic kidney disease, cancer, and inflammatory disorders, also contribute by impairing RBC production or survival. Additionally, genetic conditions such as sickle cell disease and thalassemia significantly contribute to anemia prevalence, especially in certain populations [6]. Globally, the prevalence of anemia varies significantly across regions and demographics. WHO estimates that in 2023, anemia affected 30% of women of reproductive age globally, with the highest prevalence in Africa (37.6%) and the lowest in North America (12.4%) [1]. Among children under five, the condition is even more widespread, with around 46.1% affected, particularly in low- and middle-income countries [2]. The disproportionate burden of anemia in developing regions is influenced by socioeconomic factors, healthcare access, and nutritional status [7].

Early detection of anemia is critical, particularly in vulnerable groups such as children and pregnant women, due to the serious health complications that can arise if the condition remains undiagnosed. In children, anemia can lead to significant cognitive and developmental delays, which can negatively affect academic performance and overall growth. Studies have shown that untreated anemia in children is associated with attention deficits and learning difficulties, which can hinder educational progress and social development [8]. Additionally, children in interim care often receive regular medical attention, increasing the chances of anemia screening, highlighting the importance of systematic screening across all pediatric populations [9]. In pregnant women, anemia not only affects maternal health but also poses risks to fetal development. It is linked to higher rates of preterm birth, low birth weight, and increased maternal morbidity and mortality. Ref. [10] has demonstrated that untreated anemia during pregnancy could lead to severe complications, including intrauterine growth restriction and perinatal mortality. Moreover, environmental factors, such as exposure to air pollution, can exacerbate anemia in pregnant women, emphasizing the need for targeted health interventions in high-risk areas [11]. The consequences of undiagnosed anemia extend beyond immediate health risks, as they can contribute to persistent cycles of poverty and health inequities. Vulnerable populations often encounter barriers to healthcare access, which can delay the diagnosis and treatment of anemia [12]. Therefore, implementing effective screening programs and public health initiatives aimed at early detection and management of anemia is essential. Proactively addressing anemia can improve health outcomes and the quality of life for these at-risk groups, ultimately leading to healthier communities and reducing long-term healthcare burdens.

This paper offers a comprehensive review of the application of various artificial intelligence techniques in anemia diagnosis, as discussed in previous literature. It examines key performance metrics, including accuracy, sensitivity, and specificity, for each study. By comparing different models, the paper highlights their strengths and weaknesses in terms of these metrics. Additionally, it addresses challenges such as class imbalance and data quality, which significantly affect the performance and reliability of the models.

The remaining part of this paper is presented in the following format: Section 2 reviews the application of different AI techniques in healthcare. Section 3 outlines the methodology of the whole review. Finally, Section 4 concludes by summarizing the paper's key contributions and presenting recommendations for future work.

## 2. Artificial Intelligence in Healthcare

AI has emerged as a transformative technology in healthcare, particularly in the prediction, diagnosis, and management of chronic diseases such as diabetes and cardiovascular diseases. By analyzing large amounts of medical data with advanced algorithms, AI improves diagnostic accuracy and helps create personalized treatment plans, leading to better patient outcomes. AI systems can identify patterns in complex data that may not be obvious to human doctors, resulting in earlier and more precise diagnoses.

One notable use of AI in healthcare is through ML algorithms. These algorithms are increasingly being used to estimate the risk of developing chronic conditions. For example, in the case of diabetes, ML algorithms have demonstrated significant potential in predicting the likelihood of the disease. Ref. [13] conducted a meta-analysis of ML prediction models for gestational diabetes mellitus (GDM). The study demonstrated that algorithms such as random forest (RF), extreme gradient boosting (XGBoost), and light gradient boosting machine (LightGBM) effectively identify risk factors and create early prediction models for GDM. This highlights the growing importance of ML in preventive healthcare for diabetes. Ref. [14] applied ML techniques to predict hospital readmissions for diabetic patients within 30 days. The study analyzed patient data and achieved significant predictive performance, which is crucial for improving care management and reducing healthcare costs. Ref. [15] developed a hybrid diabetes prediction model based on classification algorithms, integrating various ML techniques to enhance predictive accuracy. The study found that combining different algorithms can yield better results in identifying individuals at risk of diabetes. In Ref. [16], ML was used to assess diabetes risk based on lifestyle factors in middle-aged individuals. The study achieved high accuracy in predicting diabetes risk and emphasized the role of lifestyle modifications in diabetes prevention.

ML techniques have also been instrumental in predicting cardiovascular disease (CVD) risk factors. Ref. [17] examined the predictive ability of current ML algorithms for type 2 diabetes mellitus, which is closely linked to cardiovascular health. The study found that ML models could effectively predict the onset of diabetes, a significant risk factor for CVD. Ref. [18] discussed the integration of ML in cardiology, showing how these technologies enhance diagnostic capabilities and improve patient outcomes. The review emphasized ML's role in analyzing cardiovascular imaging data, predicting patient responses to treatments, and optimizing clinical workflows.

Recent advancements have led to the development of hybrid models that combine multiple ML techniques for improved predictive accuracy. For example, Ref. [19] proposed a Genetic Algorithm Stacking ensemble learning model to enhance diabetes risk prediction. This model can also be applied to cardiovascular risk prediction, showcasing the versatility of ML in healthcare applications. The relevance of ML in medical diagnostics is profound. ML algorithms can analyze vast amounts of medical data, including imaging, genomic, and clinical data, to identify patterns that may not be apparent to human clinicians. For instance, ML techniques have been used in medical imaging to detect tumors, fractures, and other abnormalities with high precision [20], [21]. Additionally, ML models can predict disease progression and treatment outcomes, ultimately improving patient care and resource allocation in healthcare systems [22].

The integration of ML, DL, and image processing techniques into healthcare has significantly enhanced diagnostic practices by increasing both accuracy and efficiency. These advancements are particularly noticeable in medical image analysis, where DL algorithms are highly effective at identifying patterns and anomalies that might be missed by human observers. ML has also been widely applied in biomedicine to improve diagnostic accuracy and treatment outcomes. For example, Ref. [23] explored the effectiveness of ML in predicting breast cancer therapy responses by integrating clinical, digital pathology, genomic, and transcriptomic data, showcasing the potential of advanced algorithms in multi-omics analysis. Similarly, Ref. [24] highlighted ML's use in intelligent lung nodule diagnosis, emphasizing the iterative learning ability of these algorithms to improve recognition and classification accuracy over time. This example illustrates how DL can aid in interpreting complex imaging data, contributing to more timely and accurate diagnoses. Furthermore, DL applications extend beyond traditional imaging techniques. For instance, Ref. [25] examined ML algorithms in serum profiling for kidney cancer diagnosis, achieving promising results in distinguishing between cancerous and control groups, thus underscoring the increasing role of ML in biochemical analysis for enhanced cancer diagnostics. DL's role is not limited to image analysis but also extends to processing large datasets from electronic health records (EHRs). Ref. [26] employed ML strategies to screen for gene markers associated with pulmonary arterial hypertension, demonstrating ML's capability to improve diagnostic accuracy in genetic conditions. Rapid and accurate data processing is crucial in public health settings, where timely diagnostics can be lifesaving.

Despite the many advantages, implementing these technologies in healthcare diagnostics presents several challenges. Issues such as data privacy, algorithmic bias, and the need for rigorous validation of ML models must be addressed to ensure safe, reliable, and equitable healthcare delivery [27],[28]. Proper regulatory frameworks and ethical guidelines are essential to support the widespread adoption of ML and DL technologies while maintaining patient safety and data security.

## 3. Methodology

The present state of anemia detection methods, including image processing, ML, DL, and other technological approaches, is critically analyzed in this review. Google Scholar and PubMed databases were searched for articles published between 2017 and 2024. The search strategy included terms related to image processing, ML techniques, and hybrid approaches for anemia detection. Reference lists of relevant publications were also examined to identify additional pertinent studies. The final search query used was: ("anemia detection" OR "anemia diagnosing" OR "artificial intelligence" OR "deep learning") AND ("machine learning" OR "fuzzy logic" OR "wearable devices"). Studies with insufficient data for outcome assessment, opinion or review articles, and those lacking relevant medical images or technologies were excluded. The review focuses on recent and relevant research, evaluating the application and effectiveness of various methods in anemia detection.

In this review, methods used in anemia detection are categorized into several types: Blood Analysis, Image Processing, Wearable Devices, Fuzzy Logic, ML, and Hybrid Approaches. As shown in Fig. 1, image processing is identified as the most prevalent method, comprising 40% of the approaches. Machine learning and statistical methods each account for 19%, indicating a growing trend towards data-driven techniques. Expert systems, fuzzy logic, and wearable devices are utilized in 11% of studies, highlighting their roles in decision-making and monitoring. Various techniques and models, with a focus on studies from 2017 to 2024, were analyzed. A summary of the primary research studies on ML and deep learning models, regression analysis, and other methods used in anemia detection is illustrated in Fig. 2. Common methods identified include regression techniques, ensemble methods, traditional ML approaches, deep learning models, and specialized techniques. The figure also illustrates that the Support Vector Machine (SVM) is the most frequently used technique, appearing in 10 studies, followed by K-Nearest Neighbors (KNN) in 9 studies. Other methods, such as Decision Trees and Artificial Neural Networks (ANNs), are used in 4 studies each. Ensemble methods and probabilistic techniques, including Random Forest and Naïve Bayes, are also noted. Regression techniques such as Logistic Regression and Lasso Regression are significant in statistical analysis. Table 1 provides a summary of the primary research studies on ML and deep learning models, regression analysis, and other methods used in anemia detection, highlights the diverse strategies employed to improve the accuracy and efficiency of anemia detection.

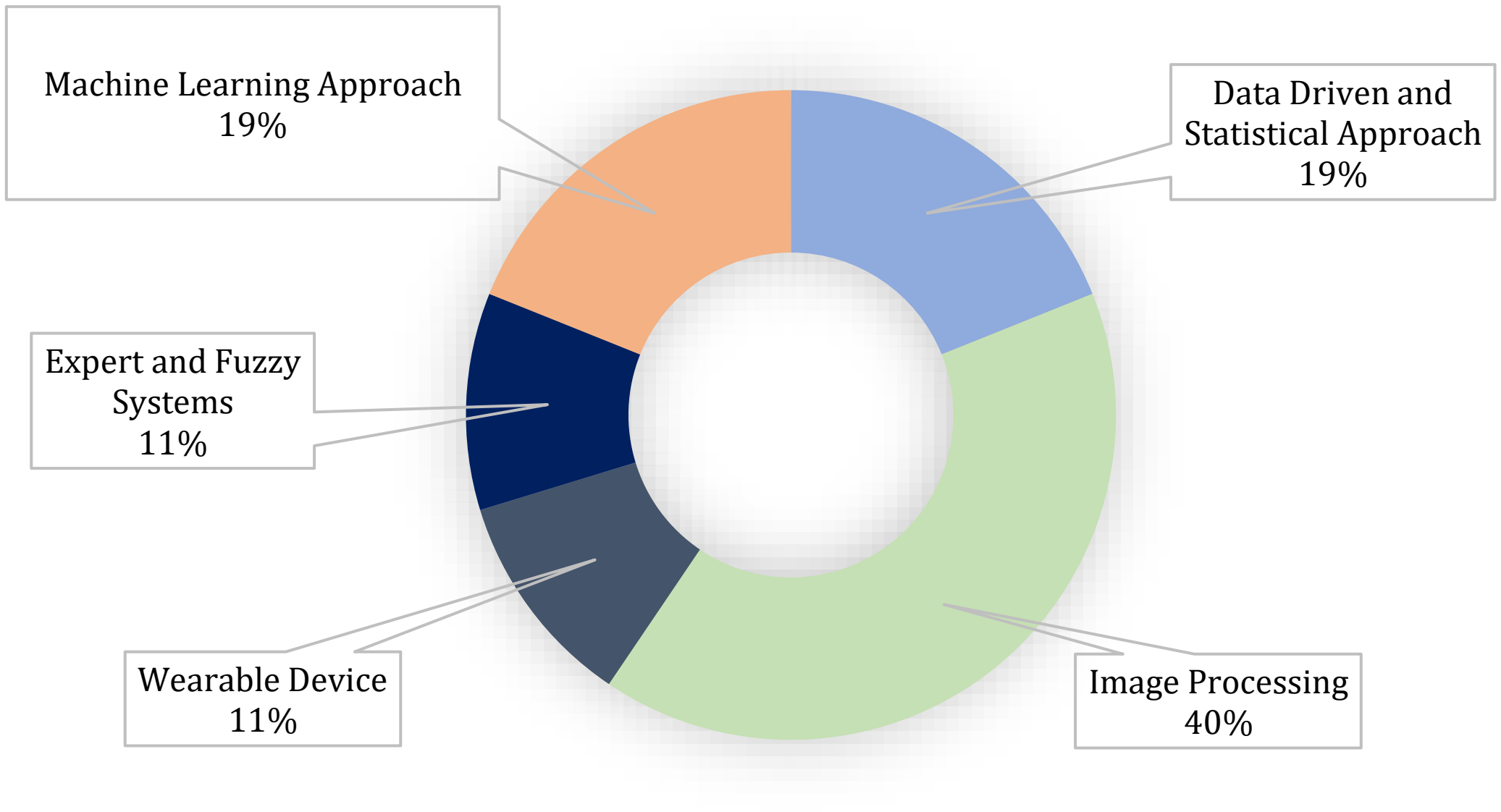

**Fig. 1:** Distribution of methodological approaches in anemia detection studies.

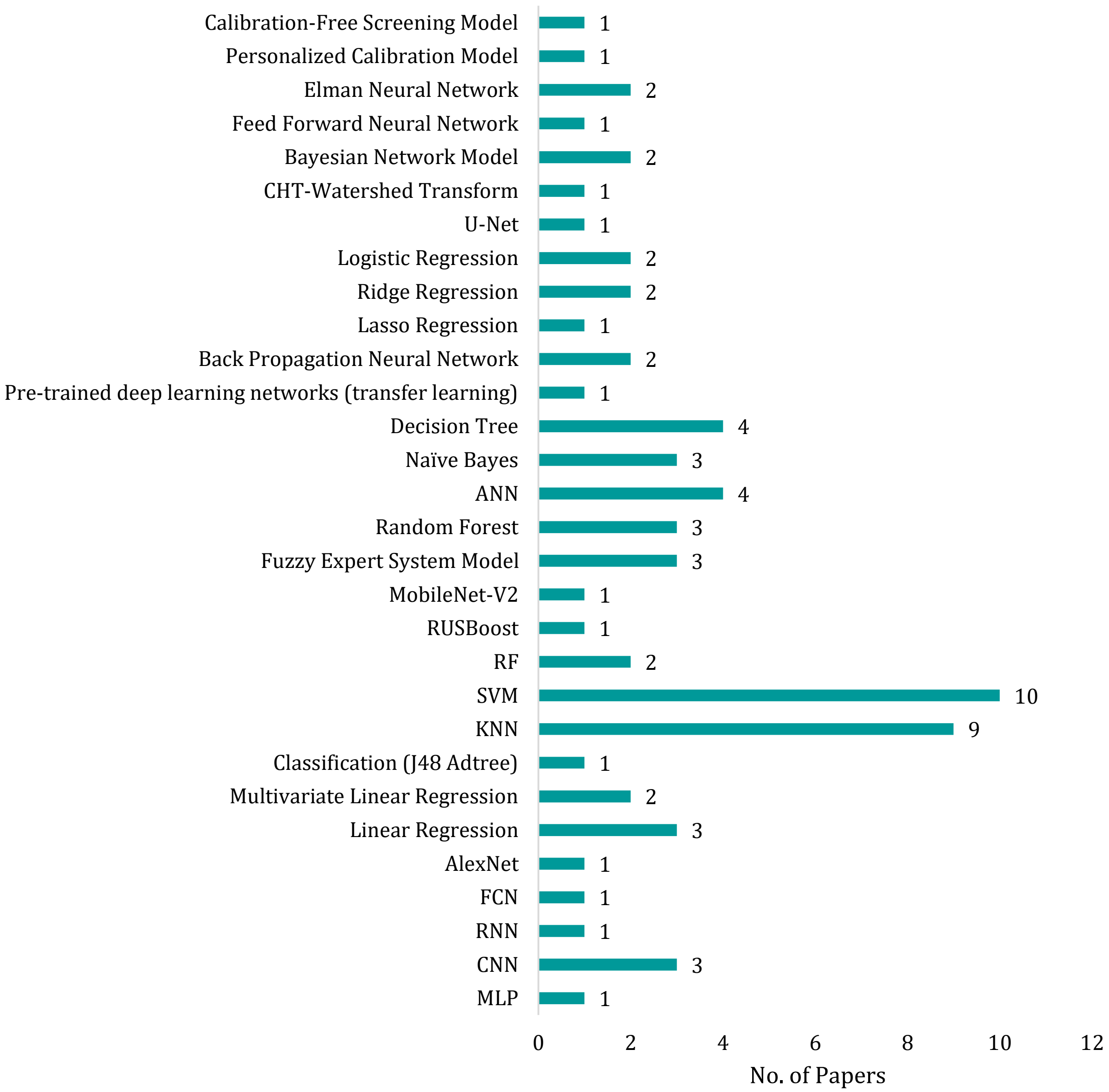

**Fig. 2:** Statistics of Methods and techniques used in reviewed literature

**Table 1:** Summary of AI models and techniques for Anemia detection: dataset, methods, and performance metrics.

| Ref. | Category | ▪ Origin<br>▪ Size of dataset | Models | Preprocessing Techniques | Performance Metrics | Accuracy | Outcomes | Limitations |
|---|---|---|---|---|---|---|---|---|
| [29] | Data driven and Statistical Approach | ▪ Database - Complete Blood Count (CBC)<br>▪ Not Specified | ▪ **MLP** (Multilayer Perceptron)<br>▪ **CNN** (Convolutional Neural Network)<br>▪ **RNN** (Recurrent Neural Network)<br>▪ **FCN** (Fully Convolutional Network) | Stratified k-Fold Cross-Validation, hyperparameter tuning | Accuracy, Precision, Recall, F1-Score, Confusion Matrix | MLP: 79%, CNN: 85%, RNN: 73%, FCN: 85% | Deep learning models showed high performance, with CNN and FCN achieving the best accuracy. Future work should focus on addressing class imbalance and refining models. | Class imbalance is a critical issue in anemia detection, where healthy samples often outnumber anemic ones. This imbalance leads to biased predictions and reduced sensitivity for the minority class. The varying performance across classes reflects inconsistent accuracy in classifying different types of anemia, which is influenced by factors such as data quality and model architecture. |
| [30] | Image Processing | ▪ Blood specimens collected from 130 sickle cell anemia (SCA) patients at Murtala Muhammad Specialist Hospital and Aminu Kano Teaching Hospital in Nigeria.<br>▪ Over 9,000 single RBC images, with each class containing 750 images. | ▪ AlexNet | Grayscale conversion, OSTU method, Hole filling | Accuracy, sensitivity, specificity, and precision calculated for the classification results. | Achieved 95.92% accuracy, 77% sensitivity, 98.82% specificity, and 90% precision. | The AlexNet model effectively classifies normal and abnormal RBCs in SCA patients, providing a robust framework for disease management. | The lack of detailed reporting on limitations restricts the understanding of potential weaknesses in the model. Additionally, lower specificity was observed due to the limited number of normal cell images in the dataset, which may affect the model's overall accuracy and reliability in clinical applications. |
| [31] | Wearable Devices | ▪ 3 different sites: Froedtert Hospital, Amader Gram (Bangladesh), Blood Center of Wisconsin.<br>▪ Transfusion study: 20, Sickle cell study: 29 participants. | ▪ Linear Regression<br>▪ Multivariate Linear Regression<br>▪ Classification (J48, ADTree) | None explicitly mentioned | Correlation coefficients, confusion matrices, ROC curves | Not applicable | The hypothesis is valid, but clinical accuracy needs improvement. Suggested enhancements include focusing on the green pixel-hemoglobin relationship and developing a custom camera app | |
| [32] | Image Processing | ▪ Not Specified<br>▪ 113 individuals' images of conjunctiva | ▪ K-Nearest Neighbors (KNN)<br>▪ Support Vector Machine (SVM) | SLIC Superpixels Algorithm | Specificity,<br>Accuracy,<br>Sensitivity | SVM: 84.4% | Demonstrate a strong correlation between the non-invasive Hb estimation and actual Hb levels | |
| [33] | Wearable Devices | ▪ Italy, India<br>▪ 218 photos of eyes | ▪ Support Vector Machine (SVM)<br>▪ K-Nearest Neighbors (KNN)<br>▪ Random Forest (RF)<br>▪ Random Under-Sampling Boost (RUSBoost)<br>▪ MobileNet Version 2 | Color standardization using LED lighting Ambient light removal | Accuracy, Sensitivity, Specificity | RUSBoost 83% | The RUSBoost algorithm, trained on palpebral conjunctiva images, shows good performance in classifying anemic and non-anemic patients. | |
| [34] | Expert and Fuzzy System | ▪ Not Specified<br>▪ 25 samples. | ▪ The fuzzy expert system model | | The study did not report any Specific accuracy metrics. However, it demonstrates the computation of the risk factor of child anemia based on the input parameters of hemoglobin and hematocrit. | Not applicable | The study showed the successful Implementation of a fuzzy expert system for assessing the risk of child anemia. The authors advocate that the fuzzy expert system approach can be extended to investigate other critical medical diseases. | The study did not provide specific accuracy metrics, making it difficult to assess the model's effectiveness. Moreover, it lacked information on the limitations of the proposed approach and the dataset used, focusing mainly on the methodology and results. This absence of critical evaluation limits understanding of the model's applicability and potential weaknesses. |
| [35] | ML Approach | ▪ Not specified, collected from an automated hematology<br>▪ 9004 records | ▪ Random Forest (RF)<br>▪ Linear Regression (LR)<br>▪ Neural Networks (NN)<br>▪ Support Vector Machines (SVM)<br>▪ Average Ensemble (AE)<br>▪ K-Nearest Neighbors Ensemble (KNN Ensemble) | Data standardization using StandardScaler, Feature selection by expert doctor, Data split into 75% training and 25% testing | Root Mean Squared Error (RMSE), Mean Absolute Error (MAE), Mean Bias Error (MBE) | KNN Ensemble achieved the lowest error | The study demonstrates the use of ML to estimate hemoglobin levels using hematological parameters. KNN Ensemble outperformed other models. | In estimating hemoglobin levels for COVID-19 patients using machine learning, the KNN Ensemble outperformed other models. However, it failed to specify the dataset source or patient demographics and did not address limitations in using machine learning for medical diagnosis. |
| [36] | ML Approach | ▪ University Hospital in Turkey,<br>▪ 1663 samples | ▪ Artificial Neural Networks (ANN) | Feature Selection Techniques | Classification Error, Area Under Curve, Precision, Recall, F-score, Accuracy | 85.6% (Bagged Decision Trees) | The system successfully classified 12 types of anemia using artificial learning methods | |
| | | | ▪ Support Vector Machines (SVM)<br>▪ Naïve Bayes (NB)<br>▪ Ensemble Decision Tree (EDT) | | | | | |
| [37] | Data driven and Statistical Approach | ▪ Blood samples from 24 subjects<br>▪ 11 healthy donors, 4 SCD, 7 HS, and 2 THAL pati 79 videos, 3442 ROIs | ▪ Pre-trained DL network (transfer learning | Automatic localization of cells in each frame - Extraction of Region of Interest (ROI) around each cell | Accuracy for healthy vs unhealthy: 91% Accuracy for healthy vs SCD, THAL, HS: 82% | Distinguishes between healthy controls and patients with an average efficiency of 91% - Distinguishes between RHHA subtypes with an efficiency of 82% | Provides a platform for in vitro monitoring of blood diseases via RBC shape analysis and combines microfluidics with deep learning for large-scale analysis of RBC deformation. | The study involved a limited number of subjects, which may affect the generalizability of the findings. Additionally, the absence of cell tracking could compromise the accuracy of the analysis over time. The potential for projection errors from the 2D visualization of a 3D scene may also impact the reliability of the results, emphasizing the need for further refinement. |
| [38] | Data driven and Statistical Approach | ▪ KEGG fatty acid pathway, Orphanet database, PubMed archive, DrugBank repository, GTEx project, GSE16334 dataset.<br>▪ 11,000 samples | ▪ HiPathia (HP)<br>▪ Random Forests (RF) | Manual curation of FA pathway, KDT gene extraction from DrugBank, gene expression data from GTEx and GSE16334 | FDR-adjusted p-value, R2 score, relevance values | Significantly different activity in REV3L and RPA complex circuits in FA patients; high predictive performance across all splits with R2 score close to 1 | Curated FA pathway demonstrates better detection of differential behavior; identified 17 genes with relevance values above 0.006, potentially influencing FA hallmarks. | Limited availability of bone marrow data in the GTEx dataset, which may restrict the comprehensiveness of the analysis. Additionally, the relatively small sample size poses a risk of overfitting, potentially undermining the generalizability of the findings. The reliance on DrugBank data for identifying drug targets may also limit the robustness of the predictions. |
| [39] | Image Processing | ▪ UK Biobank,<br>▪ 114,205 fundus images from 57,163 patients | ▪ Metadata-only<br>▪ Fundus-only | Preprocessing as described in previous work . Standardization of metadata and continuous output values. | MAE, R², AUC, Sensitivity at various specificity levels | MAE of 0.63 g/dL for Hb quantification (combined model), AUC of 0.88 for detecting anemia (combined model). | Deep learning can detect anemia and quantify Hb levels using fundus images, potentially enabling automated anemia screening. The approach is particularly useful for diabetic patients who may have regular retinal imaging. | |
| [40] | Data-Driven Approach | ▪ Not Specified<br>▪ 92 records | ▪ Linear Regression<br>▪ Multilayer Perceptron | Not Specified | Mean Absolute Percentage Error (MAPE), Mean Accuracy | Linear Regression: 11-19% MAPE for most cases; MLP: Less accurate than Linear Regression | Integral criterion can measure health state of pregnant women with anemia. Dynamics can estimate treatment effectiveness. | The data-driven approach for modeling anemia dynamics had a small dataset of only 92 records, making it sensitive to the train-test split ratio and challenging to model unique cases. |
| [41] | Fuzzy Expert System (FES) | ▪ Pathology labs and online dataset<br>▪ 150 records | ▪ Mamdani Fuzzy Inference System | Not Specified | Not Specified | 93.33% | FES accurately detects various anemia types (iron, B12, normocytic, dimorphic, polycythemia) based on HB, MCV, Iron, and B12. | The fuzzy expert system accurately detected anemia types but was limited to iron and B12 deficiency, requiring expansion to include other types. |
| [42] | Fuzzy Expert System (FES)<br>Hybrid Fuzzy-Genetic Algorithm (Fuzzy-GA) | ▪ Real data from a valid laboratory in Tehran<br>▪ 51 patients (27 healthy, 24 with microcytic anemia) | ▪ Fuzzy Expert System (FES) optimized by a Genetic Algorithm (GA) | Not explicitly mentioned in the paper, but likely involved data cleaning and normalization. | Mean Squared Error (MSE) and Accuracy | 93.6% (after optimization with Fuzzy-GA) | The proposed Fuzzy-GA system effectively diagnoses microcytic anemia with high accuracy. It combines the strengths of fuzzy expert systems and genetic algorithms, demonstrating potential for assisting physicians in early diagnosis. | The fuzzy-genetic approach for microcytic anemia diagnosis faced constraints due to its small sample size, necessitating validation with larger datasets. |
| [43] | ML Approach | ▪ The dataset was collected from various hospitals between March 1, 2017, and April 30, 2017.<br>▪ The total number of records collected was 1180, with 593 anemia positive and 588 anemia negative observations | ▪ Back-Propagation Neural Network (BPNN)<br>▪ Particle Swarm Optimization (PSO) based BP algorithm | No specific preprocessing techniques mentioned in the paper. | 1. Accuracy. Execution time | BP algorithm accuracy ranged from 53% to 72% depending on the training-testing data ratio. PSO-BP algorithm accuracy ranged from 69% to 89% depending on the training-testing data ratio. | PSO-BP algorithm performed better than the standard BP algorithm in terms of higher accuracy and lower execution time. PSO as a global optimization approach was able to train the network more effectively compared to the local optimization approach of BP. | The intelligent system using a back-propagation neural network and particle swarm optimization exhibited variable accuracy (53% to 89%) and lacked important details regarding feature engineering, potential overfitting, and external validation. |
| [44] | ML Approach | ▪ Not Specified<br>▪ Not Specified | ▪ Lasso Regression<br>▪ Ridge Regression | | Accuracy | Ridge Regression performs better than Lasso Regressio | A better and more powerful algorithm is needed to achieve maximum accuracy in medical decision making. | |
| [45] | ML Approach | ▪ The dataset was collected from different pathology centers and laboratory test centers in the nearby area.<br>▪ The dataset consists of 200 test samples. | ▪ Random Forest (RF)<br>▪ Naive Bayes (NB)<br>▪ Decision Tree (C4.5) | No specific preprocessing techniques mentioned in the paper. | The primary performance metrics used are accuracy and Mean Absolute Error (MAE). | The Naive Bayes classifier achieved the highest accuracy of 96.09%, outperforming Random Forest (95.32%) and Decision Tree C4.5 (95.46%). | The study concludes that the Naive Bayes classifier provides the best performance for predicting anemia using CBC data. | |
| [46] | ML Approach | ▪ 2011 Bangladesh Demographic and Health Survey (BDHS)<br>▪ 2013 Children under 5 years | ▪ Logistic Regression (LR)<br>▪ Linear Discriminant Analysis (LDA)<br>▪ Classification and Regression Trees (CART)<br>▪ K-Nearest Neighbors (k-NN)<br>▪ Support Vector Machines (SVM)<br>▪ Random Forest (RF) | | Accuracy, Sensitivity, Specificity, AUC (Area Under the Curve), Cohen's kappa statistic | RF algorithm achieved the best classification accuracy of 68.53% | Random Forest algorithm showed the best performance in predicting childhood anemia, Identified important features such as child morbidity, household facilities, child age, etc.ML can assist in designing targeted interventions to control anemia. | The study predicting childhood anemia in Bangladesh identified a lack of clinical and dietary variables, potential recall bias in self-reported data, and an inability to assess changes over time beyond 2011. |
| [47] | Image Processing | ▪ The dataset is from the latest public SCD RBC image dataset from MIT, collected from UPMC (University of Pittsburgh Medical Center) and MGH (Massachusetts General Hospital).<br>▪ The dataset consists of 266 raw microscopy images of 4 different SCD patients. | ▪ The proposed Deformable U-Net is compared with the standard U-Net and region growing methods. | The raw images are pre-processed by removing two-side margins and resizing them to 512 × 512 pixels. | For single-class segmentation, accuracy, precision, and F1 score are used. For multi-class segmentation, loss, accuracy, and mean IoU (Intersection over Union) are used. | For single-class segmentation, Deformable U-Net achieves 0.9960 ± 0.0003 accuracy, outperforming U-Net (0.9942 ± 0.0002) and region growing (0.9680 ± 0.0013). <br> For multi-class segmentation, Deformable U-Net achieves 0.9912 ± 0.0002 accuracy, higher than U-Net (0.9892 ± 0.0006). | The proposed Deformable U-Net framework achieves superior performance for SCD RBC semantic segmentation, particularly in handling challenges such as touching RBCs, heterogeneous shapes, blurred boundaries, and background noise. | The Deformable U-Net model for sickle cell disease RBC segmentation achieved high accuracy but was constrained by a relatively small dataset, which may limit generalizability. Additionally, its computational complexity and lack of interpretability pose challenges for practical clinical use. |
| [48] | Image Processing | ▪ Blood smear images captured under oil immersion<br>▪ Out of 233 total RBC, they found 138 normal RBC and 95 aberrant RBC | ▪ Circle Hough Transform (CHT)<br>▪ Watershed Transform | Grayscale conversion, Gaussian filtering, Canny edge detection | Number of normal and abnormal RBCs,<br>Elapsed time | Out of 233 total RBC, they found 138 normal RBC and 95 aberrant RBC using CHT in 8.710206s. But utilizing WS, it took 11.743944 seconds to distinguish between 110 aberrant and 123 normal RBCs out of 233 total. | CHT is a more efficient and robust approach for automatic diagnosis of sickle cell anemia from blood smear images. | |
| [49] | Image Processing | ▪ The dataset was collected from Al-Karama Teaching Hospital, Al-Yarmouk Teaching Hospital, and Ibn Al-Balady Hospital | ▪ Principal Component Analysis (PCA)<br>▪ Decision Tree (ID3) | Image conversion to grayscale, noise removal using median filter | Classification accuracy | The proposed algorithm achieved a best classification rate of 92% using PCA, compared to 74% without using PCA | The proposed algorithm using PCA and Decision Tree can effectively classify different types of anemia based on RBC images. | |
| [50] | Expert and Fuzzy System | ▪ Not Specified<br>▪ Not Specified | ▪ The Bayesian Network model is used to represent the causes and symptoms of hypertension and anemia. | | Not Mentioned | Not applicable | The expert system can be used to provide diagnosis and treatment recommendations for hypertension and anemia, especially in areas with limited access to medical experts. | |
| [51] | Image Processing | ▪ Tongue image dataset of healthy individuals and patients, specific size not mentioned.<br>▪ Not Specified | ▪ Support Vector Machine (SVM) for classification; features extracted include color, texture, and geometry. | Image resizing, noise removal, grayscale conversion, and edge detection (using Canny technique). | Accuracy, sensitivity, and specificity of the model; comparison with healthy individual datasets. | Achieved accuracy reported as 77.60% with specific methodologies compared (noted in related studies). | The system effectively detects diabetes and anemia through tongue analysis, indicating potential for broader applications. | Analyzing tongue images for diabetes and anemia reported a complex and time-consuming detection process, along with high costs and patient discomfort, which contributed to lower classification accuracy in some cases. |
| [52] | Image Processing | ▪ The dataset was collected by the researchers, comprising digital images of the palpebral conjunctiva.<br>▪ The original dataset had an imbalance between "safe" (no anemia risk) and "risk" (potential anemia risk) instances. The dataset was then balanced using ROSE, resulting in 127 "safe" and 127 "risk" instances. | ▪ The researchers evaluated several ML classifiers, with the kNN classifier showing the best performance. | Color clustering was applied to the images to abstract away irrelevant details. | The key performance metric was the absence of false negatives, as false negatives could lead to missed anemia cases. | With the ROSE balancing and kNN classifier, the system was able to achieve zero false negatives, ensuring reliable anemia risk detection. | The proposed system could help reduce the number of blood tests needed for anemia screening, identify suspected anemia cases, and enable widespread screening in resource-poor settings. | The computer vision system for detecting clinical signs of anemia achieved zero false negatives but did not specify limitations. It suggested that further customization based on patient characteristics, such as gender and age, could enhance performance. |
| [53] | Image Processing | ▪ The dataset was collected from various hospitals in Ghana.<br>▪ The original dataset had 527 images, which was augmented to 2,635 images using image augmentation techniques to avoid overfitting. | ▪ Naïve Bayes (NB)<br>▪ K-Nearest Neighbors (KNN)<br>▪ Convolutional Neural Network (CNN) | The images were preprocessed, but the specific techniques are not mentioned in the provided conclusion. | The models were evaluated using recall, precision, F1-score, and Area Under the Curve (AUC). | The Naive Bayes model achieved the highest accuracy at 99.96%, followed by k-NN and CNN at 99.92%, Decision Tree at 97.32%, and SVM at 94.94% | The study shows that the palpable palm is a significant and effective site for non-invasive anemia detection, surpassing previous research using the conjunctiva. This method is easier and safer, particularly for young children. The authors plan to enhance anemia detection by combining images of the palpable palm, conjunctiva, and fingernails in future work. | |
| [54] | Data Driven and Statistical Approach | ▪ Patients diagnosed with hypo-MDS and AA at the First Affiliated Hospital of Chinese Academy of Medical Sciences and the Affiliated Hospital of North China University of Science and Technology from January 1, 2008, to December 31, 2016.<br>▪ Not Specified | ▪ Logistic Regression<br>▪ Decision Trees<br>▪ Backpropagation Neural Networks (BP Neural Networks)<br>▪ Support Vector Machines (SVM) | The paper does not mention any specific preprocessing techniques used. | Sensitivity, specificity, accuracy, area under the ROC curve, Youden index, positive likelihood ratio, positive predictive value, and negative predictive value were used to evaluate the model performance. | The decision tree model had the best overall performance and was considered the optimal model for classifying hypo-MDS and AA. | The decision tree model outperformed the other models and was the optimal model for this task. Misclassification analysis provided insights into the differences between the misclassified and true positive cases. | |
| [55] | Image Processing | ▪ Two datasets are used: erythrocytesIDB1 (Dataset 1) and another unspecified dataset (Dataset 3)<br>▪ The training set size is as big as that in Scenario 4 for both datasets (1 and 3). | ▪ Model 3 has the least number of layers and filters compared to Model 1 and Model 2. Model 2 achieves the best performance. | Transfer learning and data augmentation are used to address the lack of training data. | Accuracy is the main performance metric reported. | Model 2 achieves the highest accuracy of 99.98% in Scenario 4 of Dataset 1 when using a multiclass SVM classifier. | The proposed models are efficient, lightweight, and able to extract excellent features for RBC classification. Model 2 outperforms the latest methods in the literature. | |
| [56] | Image Processing | ▪ Conjunctiva images<br>▪ Not Specified | ▪ The Bayesian Network model is used to represent the causes and symptoms of hypertension and anemia. | | Not explicitly stated, but the goal is to provide accurate diagnosis and treatment recommendations. | Not applicable | The expert system can be used to provide diagnosis and treatment recommendations for hypertension and anemia, especially in areas with limited access to medical experts. | |
| [57] | Image Processing | ▪ Anemia Detection Image Dataset<br>▪ 710 images | ▪ Naïve Bayes<br>▪ k-Nearest Neighbors (k-NN)<br>▪ Support Vector Machines (SVM)<br>▪ Convolutional Neural Networks (CNN)<br>▪ Decision Tree | The Thresholding,<br>Image Capturing,<br>ROI Extraction and Color Space Conversion,<br>Segmentation and Intensity Calculation | Accuracy Metric,<br>F1-Score Metric,<br>AUC,<br>Precision Metric,<br>Recall Metric | CNN: 99.12%<br><br>SVM: 95.4% | The palpable palm was found to be the most reliable and accurate feature for detecting iron deficiency anemia in children, with the CNN model outperforming others in accuracy, suggesting potential for future mobile app integration. | While the comparative analysis found the CNN model highly effective, it did not develop a mobile application for practical use and relied on image augmentation due to dataset scarcity. |
| [58] | Data Driven and Statistical Approach | ▪ Sejong General Hospital and Mediplex Sejong Hospital<br>▪ over 68,000 ECGs | ▪ 12-lead ECG<br>▪ 6-lead (limb lead) ECG<br>▪ Single-lead (lead I) ECG | Artifact Removal,<br>Lead Selection.<br>2D Lead Rearrangement | AUROC,<br>AUPRC,<br>Sensitivity Map Evaluation,<br>Multivariable Logistic Analysis,<br>Pearson's Correlation Coefficient | **AUROC:** 0.923(Internal Validation Dataset)<br><br>**AUROC:** 0.901(External Validation Dataset) | The DL algorithm demonstrated high effectiveness in screening for anemia using ECGs, with a focus on the QRS complex, warranting further validation through a prospective study with wearable devices. | The DL algorithm for ECGs showed lower performance with 6-lead and single-lead ECGs, necessitating further validation in real-world settings. |
| [59] | Data Driven and Statistical Approach | ▪ Gastroenterology Clinic in Derriford Hospital and Royal London Hospital<br>▪ 428 patients | ▪ ColonFlag™<br>▪ Faecal immunochemical testing (FIT) | | Sensitivity<br>Specificity<br>Positive Predictive Value (PPV)<br>Negative Predictive Value (NPV) | **FIT:**<br>Sensitivity, Specificity, PPV, NPV (reported in Table IV for colorectal cancer and high-risk adenomas at **10 µg/g cutoff**).<br>**ColonFlag:**<br>Sensitivity (100), Specificity (44), PPV (7.6), NPV (100) | The combined use of ColonFlag™ AI tool and FIT effectively prioritizes patients for colonoscopy, potentially reducing unnecessary procedures while ensuring detection of most colorectal cancer cases. | The study faced incomplete data, with critical results missing for a substantial number of patients. |
| [60] | Image Processing | ▪ Collected from participants with eye images and haemoglobin values<br>▪ 265 patients and 72 healthy subjects | ▪ Support Vector Machine (SVM)<br>▪ Artificial Neural Network (ANN)<br>▪ K-Nearest Neighbors (KNN)<br>▪ Feed-Forward Neural Network<br>▪ Elman Neural Network<br>▪ Ridge Regression | Manual Cropping of ROI, Image Rotation, Conversion to Different Color Spaces, Feature Extraction Using OpenCV | Mean Absolute Error (MAE)<br>Root Mean Square Error (RMSE)<br>Pearson's Correlation Coefficient. | Training Data:<br>Mean Absolute Error (MAE): 0.99 g/dL<br>Root Mean Squared Error (RMSE): 1.29 g/dL<br>Pearson Correlation Coefficient: 0.722<br>2 Test Data:<br>Mean Absolute Error (MAE): 1.34 g/dLRoot Mean Squared Error (RMSE): 1.72 g/dL<br>Pearson Correlation Coefficient: 0.705 | The Ridge Regression model accurately estimates hemoglobin levels from conjunctiva images, providing a convenient and accurate non-invasive anemia screening method without the need for additional hardware or calibration. | The model's performance may not generalize to diverse ethnicities and is unsuitable for individuals with abnormal conjunctival redness, limiting its applicability in certain populations. |
| [61] | Wearable Device | ▪ Collected from patients with various Anemia etiologies<br>▪ 337 subjects | ▪ Robust Multi-Linear Regression | Quality Control Algorithm,<br>Camera Flash Reflection , | Accuracy<br>Sensitivity<br>Area Under the Curve (AUC) | ±2.4 g/dL(100 subjects.) | The smartphone app for non-invasive detection of anemia using patient-sourced photos represents a significant advancement in on-demand diagnostics, | The smartphone app, while innovative, faces constraints due to testing on a single smartphone |
| | | | ▪ Personalized Calibration Model<br>▪ Calibration-Free Screening Model | Detection<br>Leukonychia Exclusion | Intraclass Correlation Coefficient (ICC)<br>Bias or Average Error | ±0.92 g/dL(16 subjects)<br>Within -2.4 g/dL(100 patients)<br>Within 0.41 g/dL (4 patients) | offering a completely non-invasive and accurate method for anemia screening and monitoring. | model, potential user error, and variability in smartphone cameras. |
| [62] | Image Processing | ▪ Maulana Azad National Institute of Technology, Bhopal (India)<br>▪ 99 Samples | ▪ Support Vector Machine (SVM) | Manual removal of poorly lit images, conversion to CIELAB color space | Accuracy, Specificity and Sensitivity | 93% accuracy was achieved | The SVM attained 93% accuracy. | The study employing SVM achieved 93% accuracy but was hindered by a limited sample size and the need for manual filtering of images for quality. This indicates a requirement for fully automated region of interest (ROI) extraction in future work. |
| [63] | Weaable Device | ▪ Sankara Nethralaya<br>▪ 200 Samples | ▪ Elman Neural System | Images captured using mobile cameras under varying lighting conditions; conversion from RGB to HSI model | Sensitivity, Specificity, Positive Predictive Value, Negative Predictive Value | The ENS achieved an accuracy of 91.3% | The Elman neural system achieved a significant result of 91.3%. | A limitation of the study is the exclusion of participants with eye diseases or recent blood donations, which may affect the generalizability of the results. Additionally, the reliance on mobile camera quality and potential variability in lighting conditions could impact the accuracy of the findings. |
| [64] | ML Approach | ▪ Dr Sardjito General Hospital in Yogyakarta, Indonesia<br>▪ 190 Samples | ▪ k-Nearest Neighbors (k-NN)<br>▪ Random Forest (RF)<br>▪ Support Vector Machine (SVM)<br>▪ Extreme Learning Machine (ELM) | Data cleansing, MinMax scaling, Label Encoding | Accuracy, Sensitivity, Specificity, Fi- Score | ELM :99.21% . | The Extreme learning machine (ELM) performed better in accuracy (99.21%) than the RF, SVM and the kNN. | A limitation is the generalizability of the Extreme Learning Machine (ELM) model, which achieved high accuracy (99.21%) but may not be applicable to broader populations or diverse clinical settings. |
| [65] | Image Processing | ▪ Italy and India,<br>▪ 99 Samples | ▪ Artificial Neural Network (ANN) | Image preprocessing to extract regions of interest (ROIs), followed by image augmentation (flipping, rotation, translation) to enhance dataset size. | Accuracy, Specificity, Sensitivity and Confusion metric | 97% accuracy was achieved. | The ANN model detected anaemia with a significant accuracy of 97%. | |

### 3.1 Evaluation of Techniques and Models for Anemia Detection

The field of anemia detection has advanced significantly, integrating both traditional laboratory methods and modern analytical techniques. In addition to these varied methodologies, it is important to consider the foundational techniques that support them. Blood analysis, which is fundamental to many detection methods, includes both traditional laboratory approaches and newer analytical techniques. Traditional blood analysis methods, such as comprehensive blood counts, remain essential but are now complemented by new technologies that improve diagnostic accuracy and accessibility. For instance, Ref. [29] compares DL architectures for anemia classification using complete blood count data, demonstrating how computational techniques can enhance conventional blood analysis. Image Processing is a significant focus, with many studies utilizing visual data from sources like retinal images or digital photographs of the conjunctiva. Ref. [32] explores non-invasive methods for estimating anemia through digital imaging. These studies leverage advanced algorithms for image segmentation and feature extraction, enabling early detection without invasive procedures.

Wearable Devices represent a growing trend in health monitoring, with studies focusing on non-invasive technologies that utilize smartphone cameras and other portable devices for anemia detection. For example, Ref. [31] highlights how these innovations make health monitoring more accessible and user-friendly, allowing individuals to conveniently track their health status. The application of Fuzzy Logic systems in predicting anemia is noteworthy. Ref. [34] and [41] illustrate how fuzzy expert systems can incorporate uncertainty and variability in patient data, offering valuable insights in clinical settings where patient presentations can vary significantly. ML is prominently featured, with a majority of studies employing various algorithms to enhance diagnostic accuracy and efficiency. Ref. [36] and [45] highlights the potential of ML techniques to classify anemia types and predict related conditions, significantly improving the diagnostic process. Hybrid Approaches combine elements from different categories to enhance diagnostic outcomes. Also, Ref. [40] utilize a data-driven approach to model the impact of control actions on anemia dynamics, illustrating how integrating various methodologies can provide a comprehensive understanding of patient health.

To detect, classify, and diagnose anemia, the investigators employed several deep learning models. According to Table 2, Artificial Neural Networks (ANNs) were used in studies [35],[36],[60] and [65]. These models are fundamental deep learning designs composed of networked neurons with advanced pattern recognition capabilities. ANNs are frequently used for regression and classification tasks due to their adaptability and flexibility. The Multi-Layer Perceptron (MLP) model was utilized in Ref. [29]. It is a specific type of feedforward neural network comprising multiple layers of neurons, including input, hidden, and output layers. MLPs are particularly effective for classification tasks because they can be trained to map input data to the correct output class. Convolutional Neural Networks (CNNs) were employed in Ref. [29], [53] and [57]. CNNs are widely recognized for their effectiveness in image processing tasks, such as detection and segmentation. They use convolutional layers to automatically extract and learn spatial properties, making them ideal for processing visual input. Recurrent Neural Networks (RNNs) were also used in Ref. [29] due to their specialization in handling sequential data. RNNs include loops to preserve context information, which makes them well-suited for applications requiring temporal patterns, such as time series analysis. Fully Convolutional Networks (FCNs) are particularly good at image segmentation tasks. Unlike regular CNNs, FCNs are designed for pixel-wise predictions, which allows for the accurate delineation of objects within images. Study [30] incorporated AlexNet, a deep CNN architecture that was pioneering in image classification. AlexNet features several convolutional and fully connected layers and demonstrated the significant potential of deep learning in visual recognition tasks. U-Net, used in study [47], is a CNN-based architecture specifically designed for biomedical image segmentation. Its encoder-decoder structure allows it to capture context and produce high-resolution segmentations, making it particularly useful for medical image analysis. The MobileNet-V2 architecture, utilized in study [33], is a lightweight CNN model optimized for mobile and edge devices. Its efficient design makes it suitable for tasks requiring a balance between accuracy and computational cost. Back Propagation Neural Networks were used in Ref. [54]. This type of neural network employs backpropagation as a training method to minimize the error rate and is commonly used for supervised learning tasks in classification and regression. The Feed Forward Neural Network (FFNN), implemented in Ref. [60], represents the simplest form of neural networks where connections between nodes do not form a cycle. FFNNs are primarily used for tasks involving straightforward input-output mappings. Studies [60] and [63] utilized the Elman Neural Network, a type of RNN that includes context layers to remember past input states. This model is particularly advantageous for tasks requiring memory of past sequences or temporal data. Additionally, the study employed pre-trained models [37], leveraging the concept of transfer learning. These networks, trained on large-scale datasets, can be fine-tuned for specific tasks such as anemia detection, offering a significant boost in performance with less computational effort and training time. The selection of each model was based on its suitability for specific uses in anemia detection, segmentation, and classification. This illustrates the wide range of applications and flexibility of deep learning techniques in medical research.

KNN is a non-parametric classification technique that categorizes new data points based on the classes of their nearest neighbors. The effectiveness of KNN in anemia detection has been demonstrated in Ref. [32], [33], [35], [46], [52], [53], [57], [60]. Conversely, SVM, a supervised learning model, constructs a hyperplane to distinguish between different classes. In addition to KNN and SVM, other ML algorithms have also been explored. Random Forest (RF), which combines multiple decision trees to improve performance, has been utilized in studies [38], [45]. Naive Bayes, a probabilistic classifier based on Bayes' theorem, has been employed in studies [36], [45], [53]. Decision Trees, which partition data based on attribute values, are used in studies [36], [45], [57], [66]. Specialized algorithms such as Bayesian Network Models [50],[56], RUSBoost [32], Fuzzy Expert System Models [34], [41], [42], CHT-Watershed Transform [48], Personalized Calibration Models [74], and Calibration-Free Screening Models [61] offer unique advantages for specific scenarios or datasets. Regression models have also played a crucial role in anemia research, offering robust frameworks for analyzing the relationships between clinical parameters and anemia indicators. Linear Regression, a foundational model, is widely used to establish initial connections between variables [67], [35], [40]. Multivariate Linear Regression, which incorporates multiple independent variables, enhances predictive accuracy for complex datasets [31]. Lasso and Ridge Regression address challenges like multicollinearity and overfitting. Lasso Regression excels in feature selection and model simplicity [44], while Ridge Regression provides stability when handling correlated predictors [44], [60]. Logistic Regression, a classification model, is commonly used to differentiate between anemic and non-anemic individuals, providing clear and interpretable results crucial for clinical decision-making [46], [54]. In the classification domain, algorithms such as J48 and ADTree have been successfully used for anemia detection. J48, an implementation of the C4.5 decision tree algorithm, is favored for its ease of use and ability to manage both categorical and numerical data, helping to differentiate between various types of anemia [31]. ADTree (Alternating Decision Tree), which incorporates boosting techniques to enhance classification performance, is particularly useful for handling complex datasets with significant variability. The combination of J48 and ADTree provides a reliable and efficient approach for classification tasks in anemia research, offering valuable insights into the factors influencing diagnosis and treatment. Table 2 presents a comparative overview of various methodologies used in this paper, including ML models, image processing techniques, wearable devices, expert systems, and data-driven approaches. Each approach is evaluated based on its summary of results, strengths, and weaknesses which highlights the trade-offs between accuracy, usability, and resource requirements, providing valuable insights for selecting appropriate strategies in clinical practice.

**Table 2:** Comparative Analysis of Approaches to Anemia Detection.

| **Detection Method** | **Performance Highlights** | **Strengths** | **Weaknesses** |
|---|---|---|---|
| ML Models | High accuracy (up to 99%), balanced sensitivity and specificity in large datasets [57]. | Robust handling of complex data; high adaptability [35], [36], [45]. | Requires significant computational resources [36]. |
| Image Processing Techniques | High accuracy (95.92% using AlexNet), high specificity (up to 98.82%), reasonable sensitivity (77%) [30], [32]. | Effective for non-invasive diagnostics[32]. | Requires high-quality images and preprocessing [30]. |
| Wearable Devices | Moderate accuracy (~83%), suitable for resource-constrained environments [33], [61] | User-friendly and accessible for frequent screening [67]. | Less precision compared to advanced models [67]. |
| Expert and Fuzzy Systems | Accuracy rates around 93.33%, effective for incomplete clinical data [34], [41]. | Handles uncertainty well; useful in absence of expert guidance [41]. | May not be as precise as advanced ML models [34]. |
| Data-driven and Statistical Approaches | Moderate accuracy (~79%), simple implementation in limited resource settings [29], [40]. | Straightforward to implement; useful for simpler cases [29]. | Less adaptable to complex data; lower accuracy [29], [40]. |

### 3.2 Limitations and Challenges in the Reviewed Literature

The limitations identified from the reviewed literature include small sample sizes, lack of detailed reporting, practical constraints, and challenges in generalizability. Addressing these issues is decisive for enhancing the effectiveness and applicability of anemia detection technologies.

Class imbalance is a significant issue for study [29], with healthy samples often outnumbering anemic ones, leading to biased predictions and reduced sensitivity for the minority class. This imbalance, combined with the lack of detailed reporting on model limitations, restricts understanding of potential weaknesses. For example, the limited number of normal cell images affected the specificity and overall reliability of some models [30]. Accuracy metrics are often missing, which hinders the assessment of model effectiveness. For instance, Ref. [35] estimating hemoglobin levels for COVID-19 patients using machine learning did not provide specific dataset sources or patient demographics, which limits the evaluation of its effectiveness. Similarly, the absence of specific accuracy metrics and details on limitations impacts the understanding of the proposed approaches in Ref. [34].

Several studies also faced constraints due to small sample sizes, which pose risks of overfitting and limit generalizability. For example, the data-driven approach for modeling anemia dynamics had only 92 records, making it sensitive to the train-test split ratio and challenging to model unique cases [40]. Similarly, the fuzzy-genetic approach for microcytic anemia diagnosis required validation with larger datasets due to its small sample size [42]. The intelligent system using a back-propagation neural network and particle swarm optimization also exhibited variable accuracy and lacked important details regarding feature engineering and potential overfitting [43]. In some cases, specific limitations were identified in the dataset and methodological approaches. For example, In Ref [47], the Deformable U-Net model for sickle cell disease RBC segmentation achieved high accuracy but was constrained by a small dataset and challenges in computational complexity and interpretability. Additionally, analyzing tongue images for diabetes and anemia reported a complex, time-consuming detection process with high costs and patient discomfort, which impacted classification accuracy [51].

However, the performance of various models and systems also showed limitations. For example, the computer vision system for detecting anemia achieved zero false negatives in Ref. [52], but lacked specified limitations and suggested that further customization based on patient characteristics could enhance performance. Similarly, while the CNN model was found highly effective in Ref. [57], it did not develop a mobile application for practical use and relied on image augmentation due to dataset scarcity. The DL algorithm for ECGs of Ref. [58] showed lower performance with 6-lead and single-lead ECGs, indicating a need for further validation in real-world settings. Other studies faced issues with incomplete data and generalizability. For instance, Ref. [59] faced incomplete data with critical results missing for a substantial number of patients. Additionally, in Ref. [60] the model's performance may not generalize to diverse ethnicities and is unsuitable for individuals with abnormal conjunctival redness. In Ref. [61], the smartphone app, while innovative, was constrained by testing on a single smartphone model, potential user error, and variability in smartphone cameras.

Finally, Ref [62] employing SVM achieved high accuracy but was hindered by a limited sample size and the need for manual image filtering, highlighting the need for fully automated region of interest (ROI) extraction in future work. Similarly, in Ref. [64] the Extreme Learning Machine (ELM) model, despite its high accuracy, may not be applicable to broader populations or diverse clinical settings. These limitations emphasize the need for further research and refinement to address these challenges and improve the robustness and applicability of anemia detection models.

## 4. Conclusions

Anemia, marked by low RBCs or hemoglobin levels, affects millions globally. Accurate diagnosis is essential for effective management and treatment. Recent advances in AI have significantly improved anemia detection, offering more precise and efficient methods. In this paper, the significant potential of AI, particularly ML and DL models, in diagnosing anemia is reviewed. Research showed that various AI techniques, including image processing, wearable devices, and expert systems, offer remarkable accuracy and usability in diverse clinical applications. SVM is the most used algorithm, with KNN closely following. Decision Tree and ANN are also prominent. Other methods like Naïve Bayes and Random Forest are used less frequently. The study emphasizes the importance of large datasets and image augmentation to enhance machine learning performance, suggesting that SVM, KNN, and Decision Tree are particularly effective for non-invasive anemia detection. Also from the studies, CNNs demonstrated high accuracy in image-based anemia detection, and wearable devices provided user-friendly solutions for frequent screening. Despite these advancements, challenges such as data quality, class imbalance, and computational complexity were noted, which could impact model generalizability. Future research should focus on expanding and diversifying datasets to include a broad range of demographics, improving class balance through techniques like data augmentation and synthetic data generation, and developing explainable AI methods to enhance model interpretability. Multi-modal approaches that integrate various data sources can offer a more comprehensive assessment of anemia. Additionally, the development of mobile and non-invasive diagnostic tools can improve accessibility, especially in resource-limited applications. Real-world validation through prospective clinical trials is essential for confirming the practical effectiveness of these models.